\documentclass[conference]{IEEEtran}
\IEEEoverridecommandlockouts
\usepackage{cite}
\usepackage{amsmath,amssymb,amsfonts}
\usepackage{algorithmic}
\usepackage{graphicx}
\usepackage{textcomp}
\usepackage{hyperref}
 \usepackage{mathtools}
 \usepackage{commath}

\usepackage{subcaption,tabularx}
\usepackage{tikz}
\def\checkmark{\tikz\fill[scale=0.4](0,.35) -- (.25,0) -- (1,.7) -- (.25,.15) -- cycle;}
\usepackage{xcolor}
\def\BibTeX{{\rm B\kern-.05em{\sc i\kern-.025em b}\kern-.08em
    T\kern-.1667em\lower.7ex\hbox{E}\kern-.125emX}}

\makeatletter
\newcommand{\linebreakand}{%
  \end{@IEEEauthorhalign}
  \hfill\mbox{}\par
  \mbox{}\hfill\begin{@IEEEauthorhalign}
}
\makeatother
\begin{document}

\title{A Graph Machine Learning Approach for Detecting Topological Patterns in Transactional Graphs}

\author{\IEEEauthorblockN{Francesco Zola, \textit{Member, IEEE}}
\IEEEauthorblockA{\textit{Digital Security,} \\
\textit{Vicomtech Foundation}\\
Donostia, Spain \\
fzola@vicomtech.org}
\and
\IEEEauthorblockN{Jon Ander Medina}
\IEEEauthorblockA{\textit{Digital Security,}\\
\textit{Vicomtech Foundation}\\
Donostia, Spain\\
jamedina@vicomtech.org}
\and
\IEEEauthorblockN{Andrea Venturi}
\IEEEauthorblockA{\textit{Digital Security,}\\
\textit{Vicomtech Foundation}\\
Donostia, Spain\\
aventuri@vicomtech.org}
  \linebreakand 
\IEEEauthorblockN{Amaia Gil}
\IEEEauthorblockA{\textit{Digital Security,}\\
\textit{Vicomtech Foundation}\\
Donostia, Spain\\
agil@vicomtech.org}
\and
\IEEEauthorblockN{Raul Orduna}
\IEEEauthorblockA{\textit{Digital Security,}\\
\textit{Vicomtech Foundation}\\
Donostia, Spain\\
rorduna@vicomtech.org}}

\maketitle

\begin{abstract}
The rise of digital ecosystems has exposed the financial sector to evolving abuse and criminal tactics that share operational knowledge and techniques both within and across different environments (fiat-based, crypto-assets, etc.). Traditional rule-based systems lack the adaptability needed to detect sophisticated or coordinated criminal behaviors (patterns), highlighting the need for strategies that analyze actors’ interactions to uncover suspicious activities and extract their \textit{modus operandi}.
For this reason, in this work, we propose an approach that integrates graph machine learning and network analysis to improve the detection of well-known topological patterns within transactional graphs. However, a key challenge lies in the limitations of traditional financial datasets, which often provide sparse, unlabeled information that is difficult to use for graph-based pattern analysis. Therefore, we firstly propose a four-step preprocessing framework that involves (i) extracting graph structures, (ii) considering data temporality to manage large node sets, (iii) detecting communities within, and (iv) applying automatic labeling strategies to generate weak ground-truth labels. Then, once the data is processed, Graph Autoencoders are implemented to distinguish among the well-known topological patterns. Specifically, three different GAE variants are implemented and compared in this analysis.
Preliminary results show that this pattern-focused, topology-driven method is effective for detecting complex financial crime schemes, offering a promising alternative to conventional rule-based detection systems.

\end{abstract}


\section{Introduction}

In the last decade, an increasing number of financial institutions have shifted from traditional, manual procedures to technology-enabled processes. This shift is driven by the digitalization of information and the automation of operations, which better position institutions to reduce bottlenecks, foster innovation, and meet regulatory requirements \cite{balkan2021impacts}. Additionally, by modernizing these processes, institutions are able to collect large volumes of customer and transaction data, which can be used to respond swiftly to evolving market dynamics, enhance customer experience and increase the operational security \cite{rodrigues2020banking}. However, to do that, financial institutions must handle and analyze their data more effectively \cite{aragani2024enhancing}. 
Therefore, they require transaction monitoring, risk assessments, and traceability tools that enhance their ability to promptly detect and address emerging risks and criminal activities within the financial ecosystem.

Legacy systems often rely on predefined rule-based models to flag suspicious activities \cite{moyes2005raise}, such as large cash deposits, frequent international transfers, or transactions involving tax havens. Although rule-based systems have proven effective to some extent \cite{gowhor2024effectiveness,gullkvist2013perceived}, they lack adaptability to change, for example, when a new \textit{modus operandi} emerges or when illicit operations are carried out through multiple transactions that do not individually raise red flags \cite{interpol}. Conversely, criminal networks have shown a high degree of adaptability, continually exploring new schemes and technologies to carry out their activities \cite{europol2025socta}, going beyond the traditional fiat-based financial ecosystem, for example, by exploiting crypto-assets \cite{chainalysis2024crypto}.


Despite the shift between fiat-based and crypto ecosystems, in many cases, criminal actors often share operational knowledge, techniques, and exploit vulnerabilities both within and across these environments \cite{calafos2022cyber}. Once a method is found to avoid detection, it is frequently replicated and adapted across other domains to evade red flags and regulatory scrutiny \cite{europol2024iocta}. These shared tactics between traditional and digital financial environments underscores the need for integrated, cross-sectoral enforcement strategies. These strategies must take into account the dynamics generated by the actor, that is, by analyzing its interactions within the considered financial networks (fiat, crypto-assets, etc.). In these cases, paradigms such as machine learning, network analysis, and graph theory must be explored to uncover concrete criminal \textit{modus operandi} and other suspicious transactional schemes. 

To address this need, we propose an approach that integrates machine learning, network analysis, and graph theory to improve the detection of suspicious operations within transactional graphs. However, access to real-world financial data is severely limited due to legal and privacy constraints. These difficulties contribute to a scarcity of high-quality datasets with reliable ground-truth labels, which restricts their use in network and graph analysis. To overcome these limitations, we introduce a generalized four-step preprocessing approach that can be applied across various financial datasets. These steps include extracting graph structures, incorporating temporal information to manage the large number of nodes, detecting communities to partition patterns, and applying automatic labeling strategies to generate weak labels as ground truth. After preprocessing, we employ \textit{Graph Machine Learning (GML)} algorithms, specifically \textit{Graph Autoencoders (GAEs)} \cite{wu2020comprehensive}, to identify well-known topological patterns extracted from both fiat-based and cryptocurrency ecosystems, which are often associated with suspicious operations. These models are designed to focus exclusively on the structural and temporal dynamics of the patterns, thereby avoiding reliance on external information such as transaction amounts, account locations, or customer profiles—factors that have already been extensively studied in the literature. In particular, three different GAE variants are trained, and their results are compared in the analysis.


The main contributions of this study are:
\begin{enumerate}
\item Presenting and validating a comprehensive methodology to extract graph-based information from financial data, based on graph creation, temporal dissection, and community detection operations;
\item Defining indicators of relevant topological structures related to suspicious activities that can be used as weak labels in an automatic process;
\item Analyzing and comparing the performance of three main GAE architectures for detecting well-known topological patterns;

\end{enumerate}

The rest of the paper is organized as follows. Section~\ref{sec:background} provides background information and reviews related work. Then, Section~\ref{sec:pattern} describes the methodology proposed in this study, while Section~\ref{sec:data} presents the datasets, the indicators, and model configurations. Then, Section~\ref{sec:results} highlights and discusses the outcomes of the analysis. Finally, conclusions and guidelines for future work are presented in Section~\ref{sec:conclusions}.

\section{Background}\label{sec:background}
This section introduces background information relevant to the proposed analysis. Specifically, Section~\ref{subsec:pattern} presents topological patterns frequently observed in suspicious activities, while Section~\ref{subsec:related} reviews approaches from related work.

\subsection{Graph Topological Patterns}\label{subsec:pattern}
In this section, we present six graph topological patterns that frequently appear in suspicious—potentially illicit—financial activities. In these transactional graphs, nodes correspond to financial accounts (senders or receivers), whereas edges are the transactions between them. These patterns are identified through a review of various bibliographic sources covering both traditional fiat-based financial systems and the cryptocurrency ecosystem \cite{dumitrescu2022anomaly, altman2023realistic,zola2024unveiling}. 

\begin{figure*}[ht]
\centering
  \begin{subfigure}[b]{0.10\linewidth}
   \includegraphics[width=\linewidth]{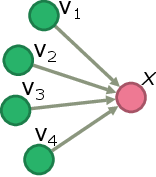}
    \caption{Collector.}
    \label{fig:collector}
  \end{subfigure}
  \hspace{0.3cm}
  \begin{subfigure}[b]{0.12\linewidth}
    \includegraphics[width=\linewidth]{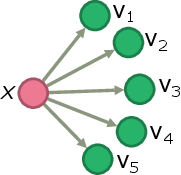}
    \caption{Sink.}
    \label{fig:sink}
  \end{subfigure}
  \hspace{0.3cm}
\begin{subfigure}[b]{0.15\linewidth}
   \includegraphics[width=\linewidth]{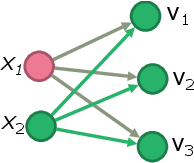}
    \caption{Collusion.}
    \label{fig:collusion}
  \end{subfigure}
      \hspace{0.3cm}
  \begin{subfigure}[b]{0.15\linewidth}
    \includegraphics[width=\linewidth]{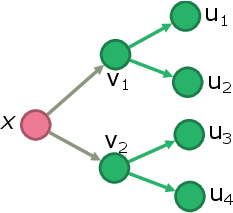}
    \caption{Branching.}
    \label{fig:branch}
  \end{subfigure}
    \hspace{0.3cm}
  \begin{subfigure}[b]{0.15\linewidth}
   \includegraphics[width=\linewidth]{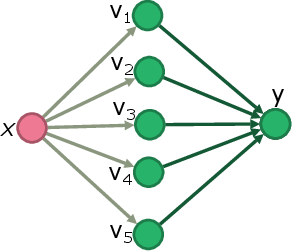}
    \caption{Scatter-Gather.}
    \label{fig:scatter}
  \end{subfigure}
  \hspace{0.2cm}
  \begin{subfigure}[b]{0.19\linewidth}
  \centering
    \includegraphics[width=\linewidth]{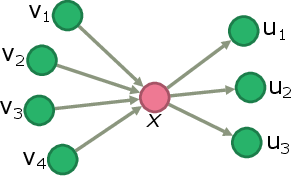}
    \caption{Gather-Scatter.}
    \label{fig:hub}
  \end{subfigure}
  \caption{Topological patterns observed in suspicious financial activities.}
  \label{fig:pattern1}
\end{figure*}

\textbf{1. Collector.} This pattern identifies a node that primarily receives funds from multiple nodes. Given a directed transactional graph $\overrightarrow{G}=(V,E)$, a node $x \in V$ represents a \textit{Collector} (or \textit{Gather}) if it receives funds from a set of nodes \{$v_1$,...,$v_n$\} $\subset V$, such that ($v_i,x$) $\in E$ for all $i=1,...,n$, with $n\geq2$, as shown in Figure \ref{fig:collector}. This pattern could potentially indicate layering mechanisms in money laundering or the presence of a dormant sender.

\textbf{2. Sink.} This pattern characterises a node that distributes funds across several recipients. Given a directed transactional graph $\overrightarrow{G}=(V,E)$, a node $x \in V$ represents a \textit{Sink} (or \textit{Scatter}) if it sends funds to a set of nodes \{$v_1$,...,$v_n$\} $\subset V$, such that ($x, v_i$) $\in E$ for all $i=1,...,n$, with $n\geq2$, as shown in Figure \ref{fig:sink}. This pattern appears in different fraud, such as smurfing or Ponzi scheme activities.

\textbf{3. Collusion.} Given a directed transactional graph $\overrightarrow{G} = (V,E)$, two (or more) nodes $x_1, x_2 \in V$ are considered to be in \textit{Collusion} if they share more than one recipient node, as shown in Figure \ref{fig:collusion}. This pattern suggests a coordinated effort between the input nodes ($x_1, x_2$), potentially aimed at concealing the origin of funds. 


\textbf{4. Branching.} Given a directed transactional graph $\overrightarrow{G}=(V,E)$, a node $x \in V$ is involved in a \textit{Branching}, if it sends money to a set of nodes \{$v_1$,...,$v_n$\} $\subset V$, with $n\geq2$, each of which subsequently transfers funds to exactly two other nodes, as shown in Figure \ref{fig:branch}. This structure forms a recursive splitting pattern, where each level of transaction results in a broader distribution of funds. Such a pattern is frequently indicative of a peeling chain, a common technique used in money laundering. 

\textbf{5. Scatter-Gather (SG).} Given a directed transactional graph $\overrightarrow{G}=(V,E)$, a node $x \in V$ represents a 
\textit{Scatter-Gather} if it sends funds to a set of nodes \{$v_1$,...,$v_n$\} $\subset V$, such that ($x,v_i$) $\in E$ for all $i=1,...,n$, with $n\geq2$. Thus, each node $v_i$ then sends funds to a common recipient $y \in V$, i.e., $(v_i, y) \in E$ for all $i = 1,...,n$ (Figure~\ref{fig:scatter}). This pattern allows the identification of cases where an input node funds a single recipient through multiple intermediaries, potentially serving as a technique to obscure the origin of the funds or evade detection in financial transactions.

\textbf{6. Gather-Scatter (GS).} Given a directed transactional graph $\overrightarrow{G}=(V,E)$, a node $x \in V$ represents a 
\textit{Gather-Scatter} if it receives funds from a set of nodes \{$v_1$,...,$v_n$\} $\subset V$, such that ($v_i,x$) $\in E$ for all $i=1,...,n$, with $n\geq2$. The node $x$ then redistributes the funds to another set of nodes \{$u_1$,...,$u_m$\} $\subset V$, such that ($x,u_i$) $\in E$ for all $i=1,...,m$, with $m\geq2$ (Figure \ref{fig:hub}). This pattern helps identify nodes that operate as a proxy, i.e., nodes that receive and send funds to a similar number of nodes.

\subsection{Related Work}\label{subsec:related}
Anti-Money Laundering (AML) and Anti-Fraud Systems (AFS) are paramount for modern financial organizations. They have rapidly evolved from manual, rule-based systems, which are unable to keep pace with the complexity and adaptiveness of fraud patterns. 
Scientific literature has therefore started to propose detection systems based on graph-based metrics to flag high-risk nodes\cite{fronzetti2017using}, community detection \cite{tariq2023topology} or based
on traditional machine learning approaches, such as ensemble classifiers, logistic regression, and anomaly detection models applied on transaction-level features \cite{chen2018machine}.
Nevertheless, these methods are typically limited to flagging anomalies based on abnormal transaction volumes or other hand-crafted indicators. This makes them both easily fooled by criminals—who can disguise illicit behavior through layering or mixing schemes—and difficult to scale in dynamic, high-volume environments.

Since financial transaction data can be naturally represented in graph form, recent research has increasingly explored graph-based machine learning as a promising paradigm \cite{dumitrescu2022anomaly, usman2023intelligent}. The rationale is that suspicious activities often manifest not only through transaction attributes but also through structural and topological patterns in the financial network. More specifically, the authors of \cite{dumitrescu2022anomaly} extract subgraphs, convert them into feature vectors, and then apply traditional anomaly detection methods, such as Isolation Forest, to identify unusual behavior. Similarly, the work in \cite{usman2023intelligent} employs Graph Convolutional Networks (GCNs) to show that graph-based methods can overcome traditional machine learning approaches in identifying suspicious accounts, even if the proposed approach require the complete account-transaction graph, resulting in a  hardly scalable workflow. 

Access to real fiat transaction datasets is heavily restricted by legal and privacy constraints. As a result, most research relies on simulators such as PaySim \cite{lopez2016paysim}, AMLWorld \cite{altman2023realistic} and AMLSim \cite{AMLSim} to generate synthetic financial transaction networks with injected money laundering behaviors. While valuable, these simulators typically provide labels at the transaction or account level, limiting their ability to capture ground-truth labels for structural patterns of suspicious activity. 

On the other hand, a large part of the scientific literature for AML based on GML focus primarily—often exclusively, on cryptocurrencies (e.g., \cite{wei2023dynamic, song2024identifying, pocher2023detecting}). For instance, \cite{wei2023dynamic} proposes a dynamic GCN model that processes sequences of transaction graphs to detect suspicious accounts using both Bitcoin-Elliptic and AMLSim data. The authors build dynamic graphs from transaction records and client information and train Graph Attention Network (GAT) based models capable of dealing with temporal information. Nevertheless, a common limitation across these works is that, despite leveraging graph representations, most still aim at classifying single accounts or transactions as suspicious (i.e., nodes or edges of the graph) \cite{shayegan2022collective,li2019survey}. However, illicit financial activities such as money laundering often emerge as entire structures of interactions (e.g., layered money flows, peeling chain, etc.) rather than isolated anomalies. For this reason, we argue that detection should focus on flagging topological structures that represent suspicious patterns rather than merely identifying individual elements. Yet, our methodology is also validated on a synthetic financial dataset (SAML-D dataset \cite{oztas2023enhancing}). However, we propose an automatic process that generates weak labels serving as ground truth for entire communities, rather than being limited to transaction-level as in most state-of-the-art studies. This enables evaluation of methods designed to detect entire topological structures of suspicious financial activity, thereby bridging the gap between node-level detection and pattern-level analysis.




\section{Method for Suspicious Pattern Detection using GML}\label{sec:pattern}

\begin{figure*}[ht]
\centering
   \includegraphics[width=0.85\linewidth]{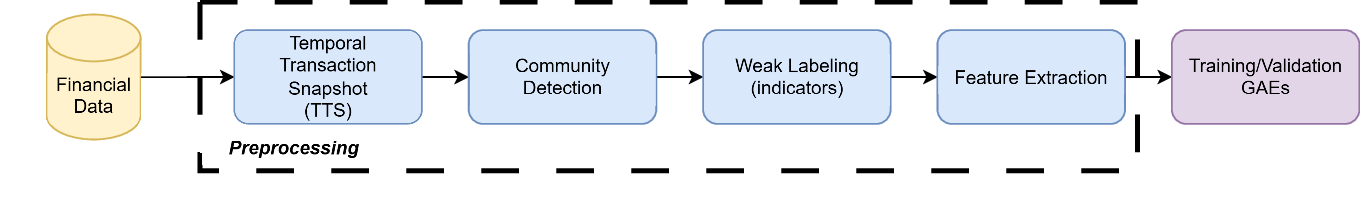}
    \caption{Schema of the proposed method for suspicious pattern detection.}
    \label{fig:phase1}
\end{figure*}

The goal of this paper is to define a comprehensive methodology based on \textit{Graph Machine Learning (GML)} models for detecting topological patterns associated with suspicious—potentially illicit—operations within transactional graphs. To this end, we employ \textit{Graph AutoEncoders (GAEs)}. The idea is to train and compare the performance of GAEs using data from a well-known financial dataset. However, financial datasets present some limitations due to their inherent time-series structure and the lack of ground-truth labels describing the patterns they can represent. To address these challenges, before training the GAE models, our approach introduces four preprocessing steps, as shown in Figure \ref{fig:phase1}.

\subsection{Preprocessing Steps}\label{subsec:preprocessing}
To build the transactional graph proposed in this work, the financial dataset should include at least two attributes: an identifier for the sender and another for the receiver. In this way, in the \textit{first step}, it is possible to leverages the intrinsic structure of these transaction data to construct a directed transactional graph $\overrightarrow{G} = (V, E)$, where nodes $V$ represent individual accounts, and directed edges $E$ represent transactions flowing from a sender to a receiver. However, using all the transactions at once to create a single static, monolithic graph can lead to several issues: the large volume of data may demand excessive computational resources and result in long graph construction times, ultimately slowing down decision-making processes. For this reason, it is essential to consider the graph's dynamic nature. To address this, a temporal dissection \cite{zola2022network} strategy is applied in this work, dividing the original dataset into fixed time intervals. The size of these time intervals is determined by a resolution parameter $\rho$. This approach generates \textit{Temporal Transaction Snapshots (TTSs)}, i.e., transactional graphs restricted to specific time intervals, designed to highlight the evolving dynamics of the graph while enhancing the scalability and usability of the overall solution. Once these TTSs are created, the \textit{second step} involves applying a community detection algorithm to identify and extract communities within the graphs, enabling their partitioning. This process transforms the general TTS into smaller, more manageable components, where each community groups together nodes that are more strongly connected to each other than to the rest of the graph. By doing so, community detection facilitates the extraction of underlying structures and recurrent patterns. Then, to evaluate the performance of the GAE models using data from the benchmark financial dataset, the extracted communities must be labeled according to the pattern they represent. 
Thus, the \textit{third step} introduces a weak labeling approach based on the definition of six indicators, each designed to validate the presence of the considered pattern. This step marks a key distinction from the state of the art, as it enables a more structured and topology-aware learning process without requiring manual annotation. Yet, in the \textit{fourth step}, node features are extracted from each labeled community and are then used to create training and validation datasets.

\subsection{Indicators for Weak Labeling}\label{subsec:indic_labeling}

In the introduced preprocessing pipeline, the indicators used to assign weak labels to communities (topological structures) play a key role. These indicators are node-level metrics computed for each node within a community, providing an additional layer of information to the graph topology. This allows for an initial separation of different patterns and supports both the training and evaluation of the GAE’s performance. In this section, we present their definitions.

\textbf{Collector:} The indicator $I_1$ (equation \ref{eq:1}) measure how closely a node $x$ resembles collector structure. This indicator takes the input degree ($deg^{-}$) of the nodes into account for the calculation. Values range from 0 to 1, with values close to 1 indicating a high probability of performing \textit{Collector} activities.

\begin{equation}\label{eq:1}
\begin{split}
    I_{1}(x) = 1- \frac{R_{1}(x)}{10*( \lfloor log_{10}(R_{1}(x))\rfloor+1 )}
     \end{split}
\end{equation}

\begin{equation}\label{eq:2}
\begin{split}
    R_{1}(x) = \left|log_{2}\left(\frac{deg^{-}(x)}{max(deg^{-}(\overrightarrow{G}))}\right)\right| \hspace{0.5cm}  deg^{-}: \text{input degree}
     \end{split}
\end{equation}

\textbf{Sink:} To measure how closely the node $x$ resembles the sink pattern, the indicator $I_2$ has been defined in equation \ref{eq:3}. This indicator is similar to $I_1$, however, in this case, the output degree ($deg^{+}$) of the nodes is taken into account during the calculation. Again, values range from 0 to 1, with values close to 1 indicating a high probability to represent a \textit{Sink}.

\begin{equation}\label{eq:3}
\begin{split}
    I_{2}(x) = 1- \frac{R_{2}(x)}{10*( \lfloor log_{10}(R_{2}(x))\rfloor+1 )}
    \end{split}
\end{equation}

\begin{equation}\label{eq:4}
\begin{split}
    R_{2}(x) = \left|log_{2}\left(\frac{deg^{+}(x)}{max(deg^{+}(\overrightarrow{G}))}\right)\right| \hspace{0.3cm}  deg^{+}: \text{output degree}
     \end{split}
\end{equation}

\textbf{Collusion:} To measure if a node $x$ is in collusions with other nodes, the indicator $I_3$ has been defined (equation \ref{eq:6}). Firstly, all the $x$'s recipient nodes are identified as $V_{out}(x)$. Then, for each $v\in V_{out}(x)$, their input nodes (i.e., the nodes that send them funds) are appended to a list $Z(v)$. For each element $z \in Z(v)$, its occurrence $\Tilde{z}$ is counted, creating the pair $(z,\Tilde{z}) \in \Tilde{Z}$. Thus, for each node that occurs more than once, its $\Tilde{z}$-value is normalized by the out-degree of $x$ (converserly, the value is excluded). This choice is made because if a node appears only once as an input, it cannot be considered relevant to the \textit{Collusion} pattern. Finally, all the normalized values are averaged as reported in equation \ref{eq:6}. This indicator ranges from 0 to 1, with values close to 1 indicating a high probability of \textit{Collusion}.






\begin{equation}\label{eq:6}
\begin{split}
    I_{3}(x) = \frac{1}{len(\Tilde{Z})}\sum_{i=1}^{len(\Tilde{Z})}\frac{\Tilde{z}_i}{deg^{+}(x)} 
\end{split}
\end{equation}



\textbf{Branching:} To measure the presence of this pattern, the indicator $I_4$ is defined in equation \ref{eq:91}, where $m$ indicates the number of nodes that receive funds from $x$, and should be $\geq$ 2. Then, $\delta$ is a parameter to establish how the number of recipients contributes to the structure. In particular, in this study, $\delta$ is set to 1. Thus, the indicator close to 1 indicates a perfect \textit{Branching}, i.e. every node has exactly two out-edges.

\begin{equation}\label{eq:91}
    I_4 (x) = \begin{cases*} 
                R_4(x) \hspace{0.3 cm} \text{if } m \geq 2\\
                0 \hspace{0.3 cm}\text{otherwise}\\
            \end{cases*}
\end{equation}

\begin{equation}\label{eq:9}
    R_4(x) = \frac{1}{m}\sum_{i=0}^{m} \delta  \hspace{1 cm}   \delta = \begin{cases*} 
                1.0 \hspace{0.3 cm} \text{if} \hspace{0.3 cm} deg^+(v_{i})= 2\\
                0 \hspace{0.3 cm}\text{otherwise}\\
            \end{cases*}
\end{equation}

\textbf{Scatter-Gather (SG):} To evaluate this pattern, the indicator $I_5$ is defined as shown in equation \ref{eq:14}. Given an initial node $x$, for each output node $y$ reachable from $x$ in two steps, the number of distinct paths from $x$ to $y$ is computed and normalized by the overall degree of the node $x$. The resulting values are then averaged. This indicator ranges from 0 to 1, with values close to 1 indicating that the intermediate nodes tend to converge toward unique recipients.

\begin{equation}\label{eq:13}
\begin{split}
        R_{5}(x) = \sum_{y_i \in Y}\frac{\# path (x, y_i)}{deg(x)} \\\text{ $Y(x)$ = set of nodes reachable in two steps}
\end{split}
\end{equation}

\begin{equation}\label{eq:14}
\begin{split}
    I_{5}(x) = \begin{cases*} 
                \frac{R_{5}(x)}{len(Y(x))} \hspace{0.3 cm} \text{if} \hspace{0.3 cm} deg^+(x)>2\\
                0 \hspace{0.3 cm}\text{otherwise}\\
            \end{cases*}
\end{split}
\end{equation}

\textbf{Gather-Scatter (GS):} To measure if a node $x$ follows a gather-scatter pattern, the indicator $I_6$ is defined in equation \ref{eq:11}. This indicator is computed by comparing the ratio between the input ($deg^-$) and output ($deg^+$) degree of the node $x$ (equation \ref{eq:10}). It is only computed when the node has a representative number of both incoming and outgoing connections, i.e., when both degrees are greater than 2. An indicator value close to 1 suggests a similar number of incoming and outgoing relations.

\begin{equation}\label{eq:10}
    R_{6}(x) = 1 - \abs{\frac{deg^+(x)-deg^-(x)}{deg^+(x)+deg^-(x)}}
\end{equation}

\begin{equation}\label{eq:11}
\begin{split}
    I_{6}(x) = \begin{cases*} 
                R_{6}(x) \hspace{0.3 cm} \text{if} \hspace{0.3 cm} deg^+(x)>2 \text{ and } deg^-(x)>2\\
                0 \hspace{0.3 cm}\text{otherwise}\\
            \end{cases*}
\end{split}
\end{equation}

A community receives a label only if at least one of its nodes has an indicator value of $\geq 0.0$. If multiple indicators meet this criterion, the community is assigned the label corresponding to the highest indicator value. This ensures that the problem remains a single-label task.


\subsection{Pattern Detection using GML}\label{subsec:autoenc}
\begin{figure}[b]
  \centering
    \includegraphics[width=0.85\linewidth]{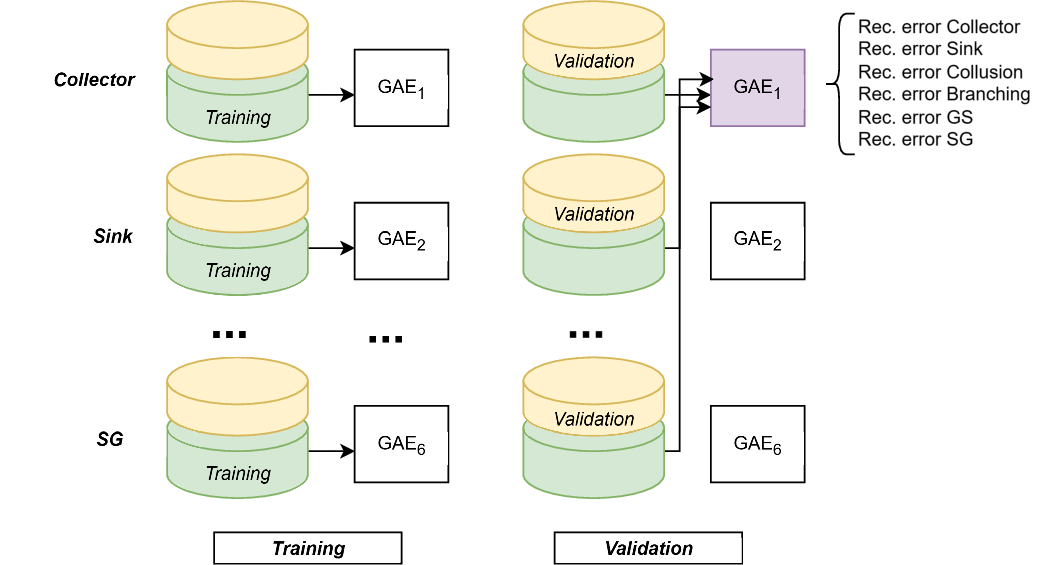}
    \caption{Training and validation schema}
    \label{fig:phase2}
\end{figure}

Once the preprocessing steps are over, the extracted weakly labeled communities are separated according to their weak labels, creating six different pattern-sets. Then, each pattern set is split in turn into training and validation sets. The first set is used for training the GAEs \cite{wu2020comprehensive}, while the second is used for evaluating their performance, as shown in Figure \ref{fig:phase2}. 

A GAE consists of two main components: an encoder and a decoder. The encoder is responsible for mapping the input graph data into a lower-dimensional latent space, while the decoder attempts to reconstruct the original input from this latent representation (embeddings) \cite{wu2020comprehensive}. In this work, for each specific pattern, a dedicated GAE is trained. As such, the model does not receive explicit labels, but instead focuses on accurately reconstructing the input samples that are separated according to their weak labels. Yet, this approach relies on the assumption that if a given pattern has been correctly learned by the model during the training stage, it should exhibit a low reconstruction error for samples of that pattern, while producing higher errors for samples of unknown patterns. Thus, as shown in Figure \ref{fig:phase2}, each GAE is evaluated not only on its corresponding validation set but also on the validation data of all other patterns, enabling a comprehensive cross-pattern assessment of generalization performance. 

The performance of three GAEs architectures based on different convolutional techniques is analyzed. Indeed, the Graph Convolutional Network (GCN) \cite{kipf2016semi}, GraphSAGE \cite{hamilton2017inductive}, and Graph Attention Network (GAT) \cite{velivckovic2017graph} are implemented in three separate model variants. From this point forward, we refer to each architecture as GAE-GCN, GAE-SAGE, and GAE-GAT, respectively. 

\section{Experimental Framework}\label{sec:data}
Section~\ref{subsec:data} introduces the dataset, while Section~\ref{subsec:tts} analyzes how the community detection and the labeling process reshape the data. Finally, Section~\ref{subsec:config} outlines the used features and the model's architectures.

\subsection{Financial Dataset}\label{subsec:data}
In this study, the dataset called \textit{SAML-D} \cite{oztas2023enhancing} is used. This dataset is designed to support the research community in evaluating anomaly detection models, benchmarking graph-based approaches, and advancing the development of more robust and interpretable AML solutions. Specifically, it contains over 9 million transactions spanning nearly one year, from October 2022 to August 2023 (11 months)\cite{oztas2023enhancing}. The dataset contains transactions from more than 800k distinct accounts.
As detailed in Section \ref{sec:pattern}, in the first step, a temporal resolution $\rho$ needs to be selected. In this study, $\rho$ is set to 7 days, generating 46 TTSs. Its implications are discussed in Section \ref{subsec:discussion}

\subsection{Community Detection and Weakly Labels}\label{subsec:tts}

In the second step, the Louvain Community (LC) detection algorithm is employed to extract communities within each TTS. LC follows a non-overlapping approach, where each node is assigned to a single community. The algorithm operates by optimising modularity, a metric that measures the density of connections within communities compared to the connections between them \cite{blondel2024fast}. Louvain is not the only viable method for community detection; indeed, alternatives such as Fast-Greedy \cite{chintalapudi2015survey}, and Girvan–Newman \cite{krishna2018analysis} have also been shown to perform well in various contexts. However, LC is chosen in this work not only because it combines scalability and computational efficiency with strong modularity optimization, but also due to its suitability for handling large-scale and complex datasets \cite{singh2020comparative}. Furthermore, its widespread adoption in the literature provides a solid basis for its application also in financial networks, ensuring methodological consistency.

\begin{figure}[ht]
\centering
    \includegraphics[width=0.85\linewidth]{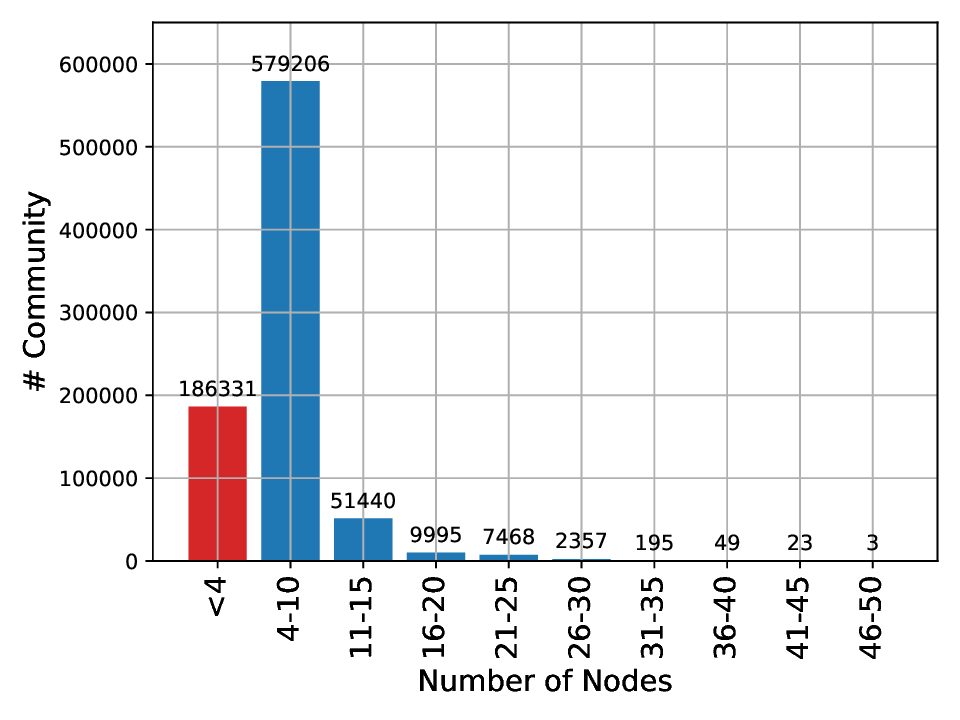}
  \caption{Number of nodes per community after the LC detection algorithm. In red are the ones excluded from the analysis.}
  \label{fig:community}
\end{figure}

The LC algorithm extracts approximately 18,000 communities from each of the TTSs, with a total of 837,067 structures identified. However, as shown in Figure \ref{fig:community}, structures with fewer than 4 nodes ($\approx$23\%) were excluded, as they did not provide sufficient information to adequately characterize the patterns under consideration. Among the remaining structures (650,736), the majority range from 4 to 10 nodes.

Then, in the third step, the indicators are computed for each community, and their distribution is reported in Table \ref{tab:testval}. Specifically, almost all available communities (650,276 of the 650,736) have at least one indicator $\geq$ 0, producing a labelled dataset with a strong class imbalance (Table \ref{tab:testval}). In fact, some patterns, such as \textit{Sink}, contain almost 600,000 samples, whereas others, such as \textit{Collusion} and \textit{Branching}, have only 14 and 112 samples, respectively. For this reason, for the minority patterns (\textit{Collusion}, \textit{Branching}, and \textit{SG}), a random oversampling (ROS) strategy \cite{mohammed2020machine} is applied only to the training sets. Yet, to construct the training set, 10,000 samples are randomly selected for each pattern, while the remaining samples are used for validation. However, for the three aforementioned minority patterns, a proportional split is applied; that is, approximately 80\% of the data is used for training and the remaining 20\% for validation. Then, on their training sets, the ROS strategy is applied to replicate the entire topological patterns until reaching 10,000 samples. By performing the split before applying ROS, we avoid introducing bias.

\begin{table}[]
\centering
\begin{tabular}{l|r|rr|c}
\hline
\multicolumn{1}{c|}{} & \textbf{\begin{tabular}[c]{@{}c@{}}Overall\\ samples\end{tabular}} & \textbf{Train} & \textbf{\begin{tabular}[c]{@{}c@{}}Train \\ \%\end{tabular}} & \textbf{\begin{tabular}[c]{@{}c@{}}ROS\end{tabular}}  \\\hline
\textit{Collector} &16,855  &10,000  &59.33  & -\\
\textit{Sink} &592,159  &10,000  &1.69 & -\\
\textit{Collusion} &14 &11 &78.57 & \checkmark\\
\textit{Branching} &112 &89 &79.46 & \checkmark\\
\textit{SG} &3,576  &2,860 &79.98 &  \checkmark  \\
\textit{GS} &37,560  &10,000 &26.62 & -\\\hline
\end{tabular}
\caption{Composition of training datasets.}
\label{tab:testval}
\end{table}

\subsection{Model Configuration}\label{subsec:config}

In this work, each decoder is implemented following the standard approach proposed in the literature, i.e., an inner product decoder \cite{kipf2016variational}, whereas the encoder architecture is modified to include three convolutional layers (\textit{CONV(32)}, \textit{CONV(16)}, and \textit{CONV(8)}), alternated with \textit{Batch Normalization}, \textit{LeakyReLU}, and \textit{Dropout} layers. As anticipated in Section \ref{subsec:autoenc}, three different convolutional strategies are implemented and compared for these layers (GCN, SAGE, and GAT), creating three distinct GAE variants. Each of these GAEs takes the adjacency matrix (representing relationships between nodes) and the node feature matrix as inputs, and attempts to reconstruct the graph structure (adjacency matrix) by minimizing the reconstruction error. In this study, the 9 metrics reported in Table \ref{tab:feaure} are used to create the node feature matrix.
GAEs are trained for a maximum of 100 epochs, with early stopping set to 3 steps, and a batch size of 25. The Adam optimisation function is used.


\begin{table}[]
\begin{tabularx}{\linewidth}{ll X}
\hline
\multicolumn{1}{c}{\textbf{\#}} & \textbf{Metric} & \multicolumn{1}{l}{\textbf{Description}}  \\\hline
\textit{1} &In-degree  &This indicates the incoming edges of a node.\\
\textit{2} &Out-degree  &This indicates the outgoing edges of a node.\\
\textit{3} &Closeness  &This centrality measures how close a node is to all other nodes in a graph \cite{wasserman1994social}  \\
\textit{4} &Betweenness  &This centrality measures how often a node lies on the shortest paths between other nodes \cite{brandes2001faster}.  \\
\textit{5} &Harmonic  & This centrality is similar to closeness centrality, but it is better suited for handling disconnected graphs.\\
\textit{6} &Second order  & This centrality measures the variability that a node is visited in a random walk \cite{kermarrec2011second}\\
\textit{7} &Laplacian  & This centrality measures how much a node contributes to the overall connectivity of the graph.    \\
\textit{8} &Constraint   & This metric measures the structural hole of the graph as shown in \cite{burt2004structural}.   \\
\textit{9} &Reciprocity   &This metric is the ratio of bidirectional edges to the total number of edges connected to a node.\\\hline
\end{tabularx}
\caption{Metrics used as node feature matrix.}
\label{tab:feaure}
\end{table}


\section{Results}\label{sec:results}

\begin{figure*}[ht]
\centering
  \begin{subfigure}[b]{0.3\linewidth}
   \includegraphics[width=\linewidth]{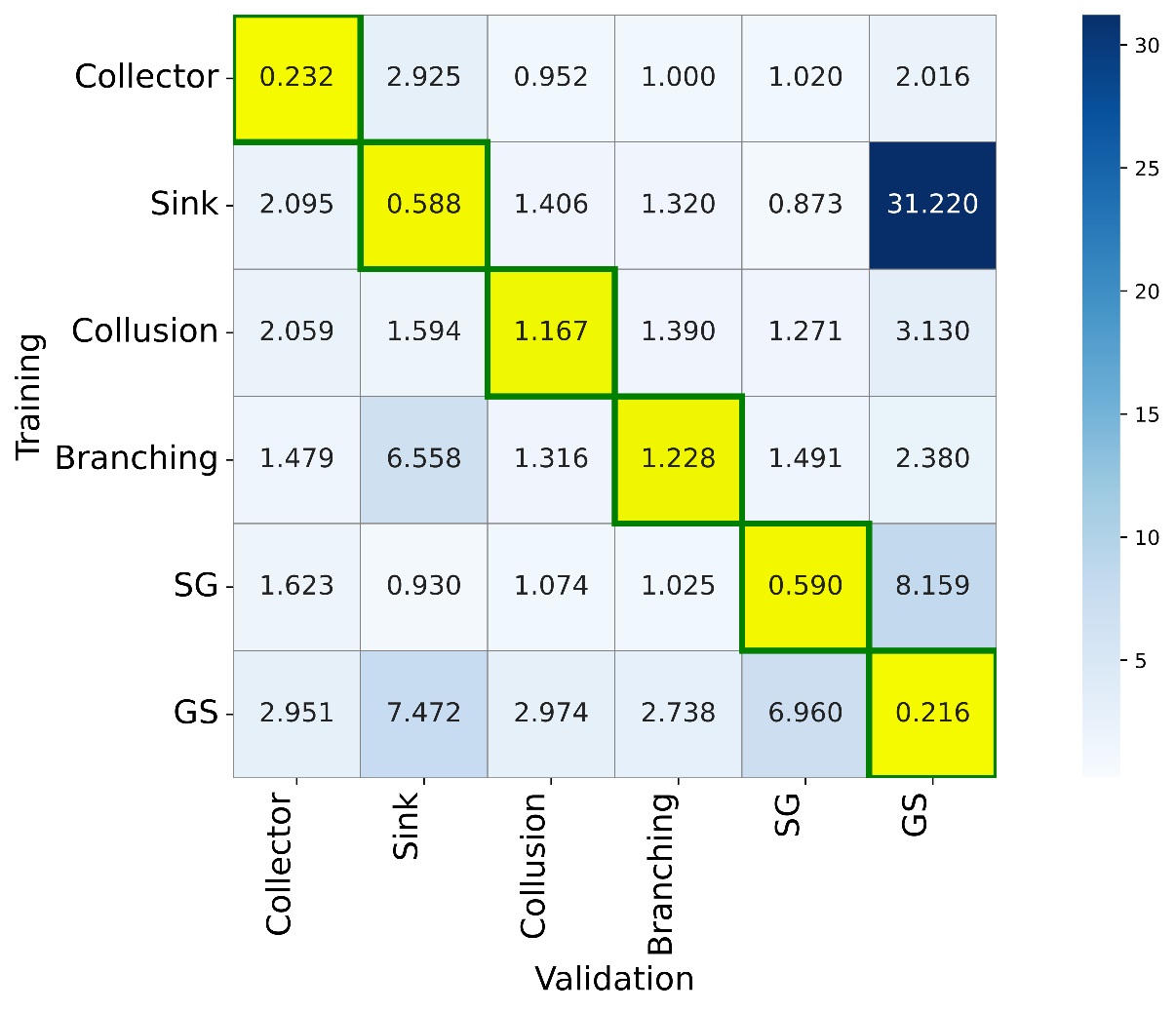}
    \caption{GAE-GCN.}
    \label{fig:gcn}
  \end{subfigure}
  \begin{subfigure}[b]{0.3\linewidth}
    \includegraphics[width=\linewidth]{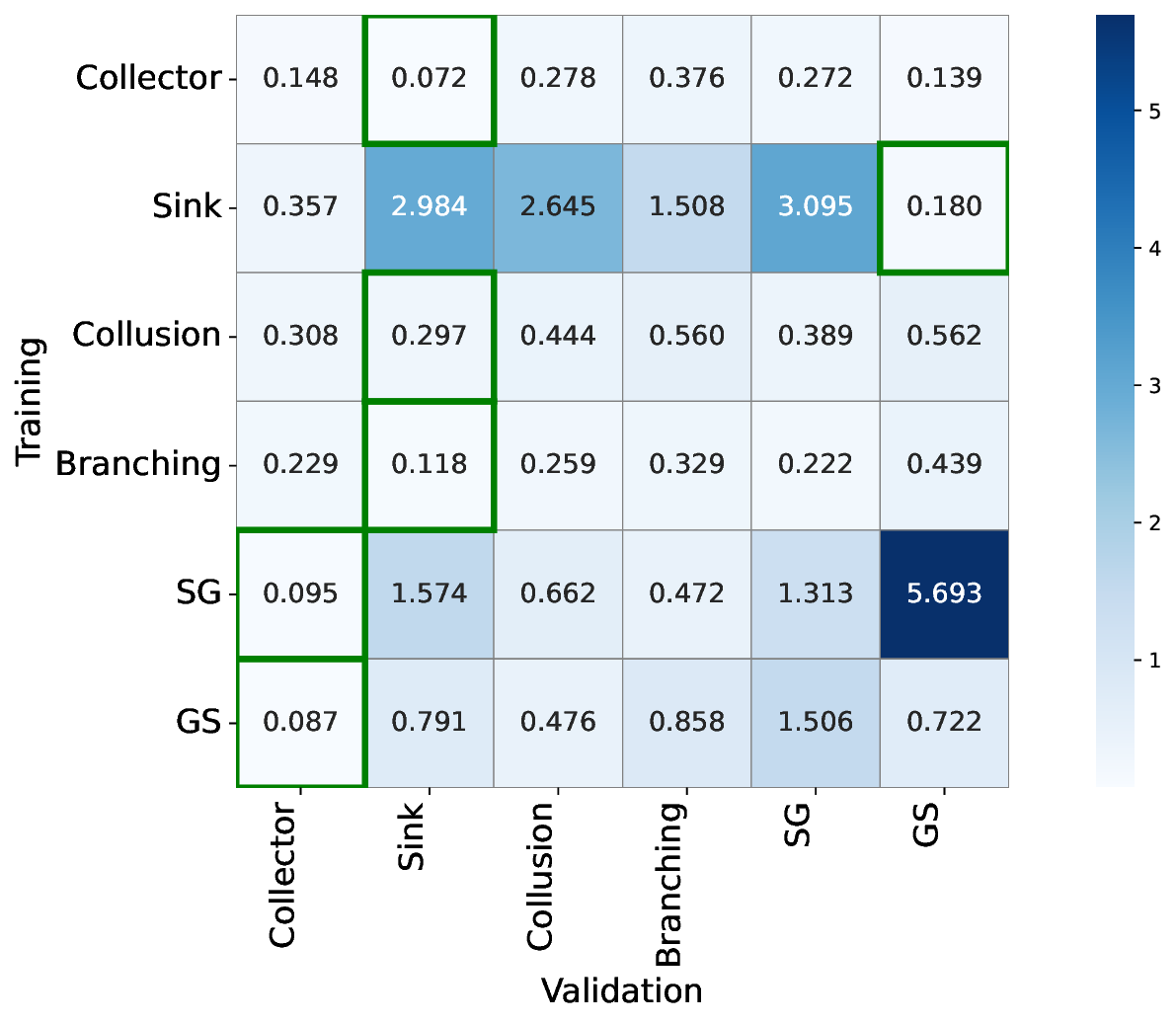}
    \caption{GAE-SAGE.}
    \label{fig:sage}
  \end{subfigure}
    \begin{subfigure}[b]{0.3\linewidth}
    \includegraphics[width=\linewidth]{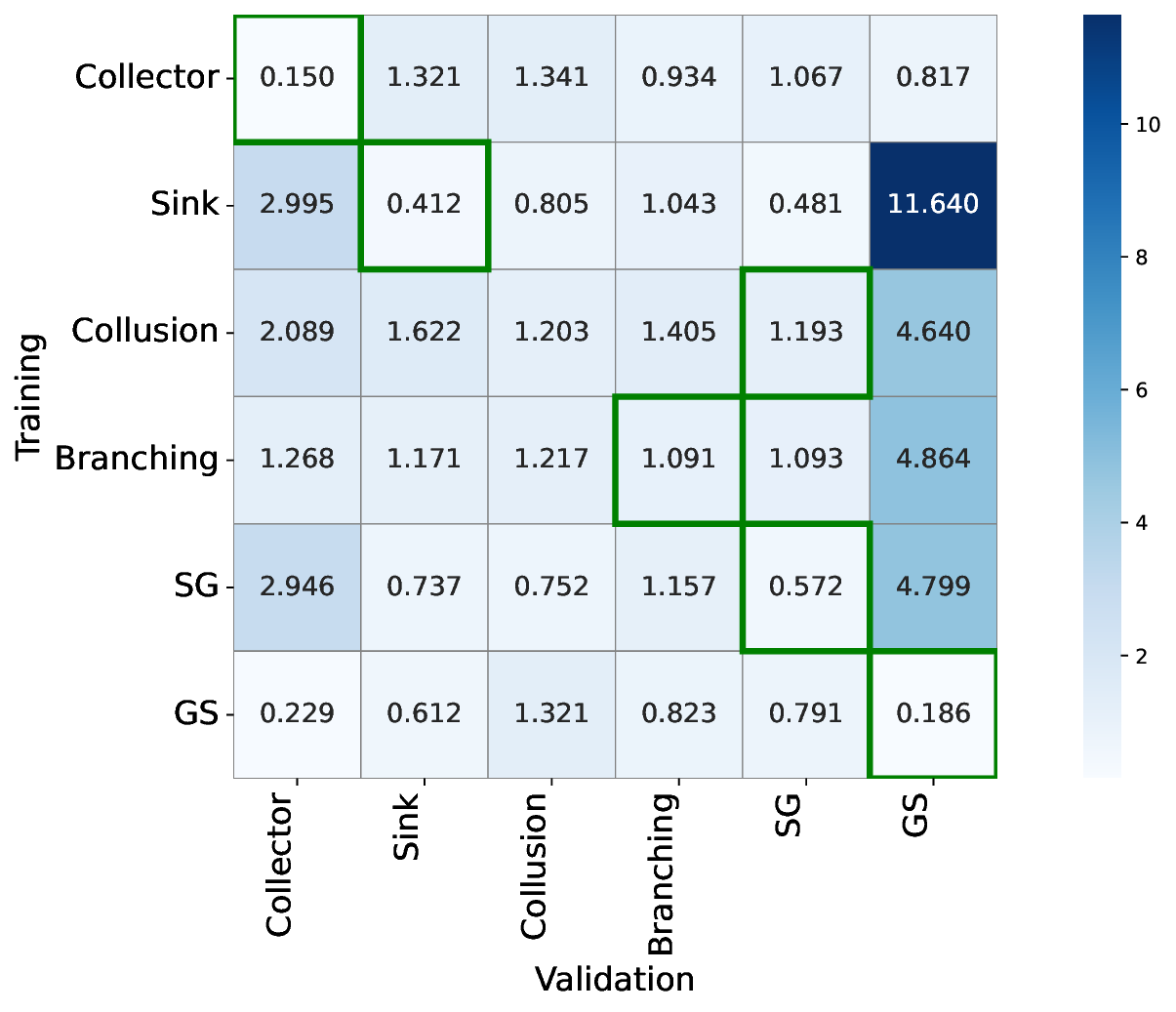}
    \caption{GAE-GAT.}
    \label{fig:gat}
  \end{subfigure}
  \caption{Reconstruction error matrix for the three GAE variants in the validation stage. The lowest error is marked with a green box, while boxes with a yellow background highlight the models with the highest degree of separability across the architectures.}
  \label{fig:loss}
\end{figure*}

The performance of the models during the validation stage is shown in Figure \ref{fig:loss}. The results clearly indicate that each GAE-GCN model, when trained on a specific pattern, is able to detect that pattern with the lowest reconstruction error compared to the other patterns. That is, the lowest reconstruction errors are on the diagonal of the reported matrix, as shown in Figure \ref{fig:gcn}. Similarly, GAE-GAT models correctly detect 5 out of the 6 patterns, failing only on the model trained with the \textit{Collusion} pattern, which is confused with the \textit{SG} pattern. Finally, GAE-SAGE models, in their current implementations, are unable to detect any pattern correctly.

To identify the best overall model across all architectures, the model that achieves the highest degree of separability is considered for each pattern. That is, among the models with the lowest reconstruction error on the training pattern, those that also exhibit the highest reconstruction errors on all other patterns are chosen.
In fact, although GAE-GAT achieves the lowest reconstruction error in the 5 correctly detected patterns compared to the GAE-GCN, the latter shows the highest degree of separability, emerging as the best solution across all patterns (yellow model in Figure \ref{fig:loss}).

\subsection{Discussion and Limitations}\label{subsec:discussion}

This analysis demonstrates that GAEs provide a promising approach for learning and detecting topological patterns commonly observed in suspicious financial activities. Although some implementations, such as GAE-SAGE, require substantial improvements due to poor performance, others, such as GAE-GCN models (the best-performing architectures), achieve the lowest reconstruction error for the patterns on which they were trained. This property can be leveraged to determine whether a new sample matches a specific trained pattern, enabling their application in real-world AML systems.

However, it is important to acknowledge several limitations of this analysis. The financial dataset used in this study inherently introduces constraints due to its construction. As previously discussed, it lacks explicit structural labels that would typically describe the underlying patterns within the data, which limits our ability to perform a fully comprehensive evaluation of the approach. Although an automatic labeling process based on defined indicators was introduced to mitigate this shortcoming, it detected only a few samples for patterns such as \textit{Collusion} and \textit{Branching}. This raises concerns about severe class imbalance and its impact on the quality of the trained machine learning models. While a basic oversampling technique (ROS) was applied, further advanced techniques could improve model performance. Finally, another limitation arises from the temporal resolution selected ($\rho$). Lower values could produce graphs with too few nodes to adequately represent the patterns of interest, while higher values could reduce applicability in real-world scenarios where timing is critical. Therefore, varying the choice of $\rho$ may influence the results obtained in this analysis.




\section{Conclusion and Future Work}\label{sec:conclusions}

This work combines machine learning, network analysis, and graph theory to improve the detection of topological patterns commonly associated with suspicious operations in transactional graphs. Specifically, six patterns that capture structures shared across multiple economic systems and reflect behaviors characteristic of large-scale transaction networks are analyzed. The work starts by introducing and validating a generalized preprocessing pipeline that addresses the lack of structural (pattern) and high-quality ground-truth information in financial datasets. Then, the application of graph autoencoders, especially GAE-GCN, demonstrates the potential of focusing solely on topological and temporal dynamics, offering a scalable and data-efficient alternative to traditional methods that rely heavily on external attributes. This multi-phase approach constitutes an initial step toward scalable and interpretable methods for detecting suspicious structures in large transactional networks.

Future work should focus on applying solutions that address the imbalance problem and provide more variability on the generated samples, such as Generative Adversarial Networks (GANs) \cite{zola2022network} or graph sampling techniques \cite{leskovec2006sampling}. Finally, incorporating analyst feedback into the model training process through human-in-the-loop strategies can improve model adaptability and interpretability. Yet, analysts possess domain expertise that can help prioritize relevant features, validate emerging patterns, and label ambiguous cases that automated systems may misclassify. In this way, we aim to support financial analysts in identifying suspicious—potentially illicit—activities with greater speed and precision, contributing to more proactive and resilient fraud mitigation strategies.
%

\section*{Acknowledgment}
This work has been partially supported by the European Union's Horizon 2020 Research and Innovation Program under the project CEDAR (Grant Agreement No. 101135577) and FALCON (Grant Agreement No. 101121281). The content of this article does not reflect the official opinion of the European Union. Responsibility for the information and views expressed therein lies entirely with the authors.

\bibliographystyle{splncs04}
\bibliography{main}

\begin{thebibliography}{10}
\providecommand{\url}[1]{\texttt{#1}}
\providecommand{\urlprefix}{URL }
\providecommand{\doi}[1]{https://doi.org/#1}

\bibitem{altman2023realistic}
Altman, E., Blanu{\v{s}}a, J., Von~Niederh{\"a}usern, L., Egressy, B., Anghel,
  A., Atasu, K.: Realistic synthetic financial transactions for anti-money
  laundering models. Advances in Neural Information Processing Systems
  \textbf{36},  29851--29874 (2023)

\bibitem{aragani2024enhancing}
Aragani, V.M., Maroju, P.K., Mudunuri, L.N.R.: Enhancing cybersecurity in
  banking: Best practices and solutions for securing the digital supply chain.
  Journal of Computational Analysis \& Applications  (2024)

\bibitem{balkan2021impacts}
Balkan, B.: Impacts of digitalization on banks and banking. In: The Impact of
  Artificial Intelligence on Governance, Economics and Finance, Volume I, pp.
  33--50. Springer (2021)

\bibitem{blondel2024fast}
Blondel, V., Guillaume, J.L., Lambiotte, R.: Fast unfolding of communities in
  large networks: 15 years later. Journal of Statistical Mechanics: Theory and
  Experiment  \textbf{2024}(10),  10R001 (2024)

\bibitem{brandes2001faster}
Brandes, U.: A faster algorithm for betweenness centrality. Journal of
  mathematical sociology  \textbf{25}(2),  163--177 (2001)

\bibitem{burt2004structural}
Burt, R.S.: Structural holes and good ideas. American journal of sociology
  \textbf{110}(2),  349--399 (2004)

\bibitem{calafos2022cyber}
Calafos, M.W., Dimitoglou, G.: Cyber laundering: Money laundering from fiat
  money to cryptocurrency. In: Principles and Practice of Blockchains, pp.
  271--300. Springer (2022)

\bibitem{chainalysis2024crypto}
{Chainalysis Inc.}: {The 2024 Crypto Crime Report} (2024),
  \url{https://go.chainalysis.com/crypto-crime-2024.html}, accessed on
  06/08/2025

\bibitem{chen2018machine}
Chen, Z., Van~Khoa, L.D., Teoh, E.N., Nazir, A., Karuppiah, E.K., Lam, K.S.:
  Machine learning techniques for anti-money laundering (aml) solutions in
  suspicious transaction detection: a review. Knowledge and Information Systems
   \textbf{57}(2),  245--285 (2018)

\bibitem{chintalapudi2015survey}
Chintalapudi, S.R., Prasad, M.K.: A survey on community detection algorithms in
  large scale real world networks. In: 2015 2nd international conference on
  computing for sustainable global development (INDIACom). pp. 1323--1327. IEEE
  (2015)

\bibitem{dumitrescu2022anomaly}
Dumitrescu, B., B{\u{a}}ltoiu, A., Budulan, {\c{S}}.: Anomaly detection in
  graphs of bank transactions for anti money laundering applications. IEEE
  Access  \textbf{10},  47699--47714 (2022)

\bibitem{europol2024iocta}
Europol: Iocta 2024. Internet Organised Crime Threat Assessment  (2024),
  \url{https://www.europol.europa.eu/cms/sites/default/files/documents/Internet%20Organised%20Crime%20Threat%20Assessment%20IOCTA%202024.pdf},
  accessed on 06/08/2025

\bibitem{europol2025socta}
Europol: Eu-socta 2025. Serious and Organised Crime Threat Assessment  (2025),
  \url{https://www.europol.europa.eu/cms/sites/default/files/documents/EU-SOCTA-2025.pdf},
  accessed on 06/08/2025

\bibitem{fronzetti2017using}
Fronzetti~Colladon, A., Remondi, E.: Using social network analysis to prevent
  money laundering. Expert Systems with Applications  \textbf{67},  49--58
  (2017). \doi{https://doi.org/10.1016/j.eswa.2016.09.029}

\bibitem{gowhor2024effectiveness}
Gowhor, H.S.: Effectiveness criteria of suspicious transaction reporting for
  early detection of terrorist financing activities. Journal of Financial Crime
   \textbf{31}(6),  1516--1531 (2024)

\bibitem{gullkvist2013perceived}
Gullkvist, B., Jokipii, A.: Perceived importance of red flags across fraud
  types. Critical Perspectives on Accounting  \textbf{24}(1),  44--61 (2013)

\bibitem{hamilton2017inductive}
Hamilton, W., Ying, Z., Leskovec, J.: Inductive representation learning on
  large graphs. Advances in neural information processing systems  (2017)

\bibitem{interpol}
Interpol: Interpol financial fraud assessment. INTERPOL Financial Fraud
  assessment  (2024), accessed on 06/08/2025

\bibitem{kermarrec2011second}
Kermarrec, A.M., Le~Merrer, E., Sericola, B., Tr{\'e}dan, G.: Second order
  centrality: Distributed assessment of nodes criticity in complex networks.
  Computer Communications  \textbf{34}(5),  619--628 (2011)

\bibitem{kipf2016variational}
Kipf, T.N., Welling, M.: Variational graph auto-encoders. arXiv preprint
  arXiv:1611.07308  (2016)

\bibitem{kipf2016semi}
Kipf, T.: Semi-supervised classification with graph convolutional networks.
  arXiv preprint arXiv:1609.02907  (2016)

\bibitem{krishna2018analysis}
Krishna, R.J., Chaudhry, Y., Sharma, D.P.: Analysis of community detection
  algorithms. In: 2018 second international conference on inventive
  communication and computational technologies (ICICCT). pp. 669--674. IEEE
  (2018)

\bibitem{leskovec2006sampling}
Leskovec, J., Faloutsos, C.: Sampling from large graphs. In: Proceedings of the
  12th ACM SIGKDD international conference on Knowledge discovery and data
  mining. pp. 631--636 (2006)

\bibitem{li2019survey}
Li, J., Gu, C., Wei, F., Chen, X.: A survey on blockchain anomaly detection
  using data mining techniques. In: International Conference on Blockchain and
  Trustworthy Systems. pp. 491--504. Springer (2019)

\bibitem{lopez2016paysim}
Lopez-Rojas, E., Elmir, A., Axelsson, S.: Paysim: A financial mobile money
  simulator for fraud detection. In: 28th European Modeling and Simulation
  Symposium, EMSS, Larnaca. pp. 249--255. Dime University of Genoa (2016)

\bibitem{mohammed2020machine}
Mohammed, R., Rawashdeh, J., Abdullah, M.: Machine learning with oversampling
  and undersampling techniques: overview study and experimental results. In:
  2020 11th international conference on information and communication systems
  (ICICS). pp. 243--248. IEEE (2020)

\bibitem{moyes2005raise}
Moyes, G.D., Lin, P., Landry~Jr, R.M.: Raise the red flag: a recent study
  examines which sas no. 99 indicators are more effective in detecting
  fraudulent financial reporting. Internal Auditor  \textbf{62}(5),  47--52
  (2005)

\bibitem{oztas2023enhancing}
Oztas, B., Cetinkaya, D., Adedoyin, F., Budka, M., Dogan, H., Aksu, G.:
  Enhancing anti-money laundering: Development of a synthetic transaction
  monitoring dataset. In: 2023 IEEE International Conference on e-Business
  Engineering (ICEBE). pp. 47--54. IEEE (2023)

\bibitem{pocher2023detecting}
Pocher, N., Zichichi, M., Merizzi, F., Shafiq, M.Z., Ferretti, S.: Detecting
  anomalous cryptocurrency transactions: An aml/cft application of machine
  learning-based forensics. Electronic Markets  \textbf{33}(1), ~37 (2023)

\bibitem{rodrigues2020banking}
Rodrigues, J.F., Ferreira, F.A., Pereira, L.F., Carayannis, E.G., Ferreira,
  J.J.: Banking digitalization:(re) thinking strategies and trends using
  problem structuring methods. IEEE Transactions on Engineering Management
  \textbf{69}(4),  1517--1531 (2020)

\bibitem{shayegan2022collective}
Shayegan, M.J., Sabor, H.R., Uddin, M., Chen, C.L.: A collective anomaly
  detection technique to detect crypto wallet frauds on bitcoin network.
  Symmetry  \textbf{14}(2), ~328 (2022)

\bibitem{singh2020comparative}
Singh, D., Garg, R.: Comparative analysis of sequential community detection
  algorithms based on internal and external quality measure. Journal of
  Statistics and Management Systems  \textbf{23}(7),  1129--1146 (2020)

\bibitem{song2024identifying}
Song, K., Dhraief, M.A., Xu, M., Cai, L., Chen, X., Mithal, A., Chen, J.:
  Identifying money laundering subgraphs on the blockchain. In: Proceedings of
  the 5th ACM International Conference on AI in Finance. pp. 195--203 (2024)

\bibitem{tariq2023topology}
Tariq, H., Hassani, M.: Topology-agnostic detection of temporal money
  laundering flows in billion-scale transactions. In: Joint European Conference
  on Machine Learning and Knowledge Discovery in Databases. pp. 402--419.
  Springer (2023)

\bibitem{AMLSim}
Toyotaro, S., Hiroki, K.: {Anti-Money Laundering Datasets}: {InPlusLab}
  anti-money laundering datadatasets. \url{http://github.com/IBM/AMLSim/}
  (2021)

\bibitem{usman2023intelligent}
Usman, A., Naveed, N., Munawar, S.: Intelligent anti-money laundering fraud
  control using graph-based machine learning model for the financial domain.
  Journal of Cases on Information Technology (JCIT)  (2023)

\bibitem{velivckovic2017graph}
Veli{\v{c}}kovi{\'c}, P., Cucurull, G., Casanova, A., Romero, A., Lio, P.,
  Bengio, Y.: Graph attention networks. arXiv preprint arXiv:1710.10903  (2017)

\bibitem{wasserman1994social}
Wasserman, S.: Social network analysis: Methods and applications. The Press
  Syndicate of the University of Cambridge  (1994)

\bibitem{wei2023dynamic}
Wei, T., Zeng, B., Guo, W., Guo, Z., Tu, S., Xu, L.: A dynamic graph
  convolutional network for anti-money laundering. In: International Conference
  on Intelligent Computing. pp. 493--502. Springer (2023)

\bibitem{wu2020comprehensive}
Wu, Z., Pan, S., Chen, F., Long, G., Zhang, C., Yu, P.S.: A comprehensive
  survey on graph neural networks. IEEE transactions on neural networks and
  learning systems  \textbf{32}(1),  4--24 (2020)

\bibitem{zola2024unveiling}
Zola, F., Gorricho, M., Medina, J.A., Segurola, L., Orduna-Urrutia, R.:
  Unveiling dynamics and patterns: A comprehensive analysis of spreading
  patterns and similarities in low-labelled ransomware families. In: 2024 IEEE
  International Conference on Blockchain (Blockchain). pp. 260--268. IEEE
  (2024)

\bibitem{zola2022network}
Zola, F., Segurola-Gil, L., Bruse, J.L., Galar, M., Orduna-Urrutia, R.: Network
  traffic analysis through node behaviour classification: a graph-based
  approach with temporal dissection and data-level preprocessing. Computers \&
  Security  \textbf{115},  102632 (2022)

\end{thebibliography}



\end{document}